\documentclass{article}
\usepackage{authblk}
\usepackage[utf8]{inputenc}
\usepackage{multirow}
\usepackage{adjustbox}
\usepackage{float}
\usepackage{url}
\usepackage{array}

\usepackage[english]{babel}  

\providecommand{\keywords}[1]
{
  \small	
  \textbf{\textit{Keywords:}} #1
}

\usepackage{graphicx}

\usepackage{epstopdf} 

\addto\captionsenglish{%
}

\title{Evaluating the Performance of Large Language Models for Spanish Language in Undergraduate Admissions Exams}

\author[1]{Sabino Miranda}
\author[2]{Obdulia Pichardo-Lagunas}
\author[2]{Bella Martínez-Seis}
\author[3]{Pierre Baldi}

\affil[1]{IEEE member, Mexico City, Mexico }
\affil[2]{UPIITA-IPN, Mexico City, Mexico }
\affil[3]{University of California, Irvine, CA, USA}             
\affil[ ]{smiranda@ieee.org, \{opichardola, bcmartinez\}@ipn.mx, pfbaldi@ics.uci.edu}
\date{} 
\begin{document}

\maketitle

\renewcommand{\tablename}{Table}
\begin{abstract}
This study evaluates the performance of large language models, specifically GPT-3.5 and BARD (supported by Gemini Pro model), in undergraduate admissions exams proposed by the National Polytechnic Institute in Mexico. The exams cover Engineering/Mathematical and Physical Sciences, Biological and Medical Sciences, and Social and Administrative Sciences. Both models demonstrated proficiency, exceeding the minimum acceptance scores for respective academic programs to up to 75\% for some academic programs. GPT-3.5 outperformed BARD in Mathematics and Physics, while BARD performed better in History and questions related to factual information. Overall, GPT-3.5 marginally surpassed BARD with scores of 60.94\% and 60.42\%, respectively.
\end{abstract}







\section{Introduction}
\label{sec:introduction}

In recent years, the landscape of education has been significantly influenced by the remarkable advancements in generative artificial intelligence and large language models (LLMs). These innovations have paved the way for many educational technology solutions, aiming to streamline the often cumbersome and time-consuming tasks associated with generating and analyzing textual content. These models, exemplified by Generative Pre-trained Transformer (GPT) \cite{openai2023gpt4}, harness deep learning, reinforcement learning, and self-attention mechanisms to process and generate human-like text based on natural language inputs. Their capability to comprehend intricate patterns and relationships within textual content, encompassing semantic, contextual, and syntactic nuances, has revolutionized various sectors, including education \cite{10.3389/feduc.2023.1206936}, \cite{deWinter2023}, \cite{fi15060192}. 

LLMs such as GPT-3.5 \cite{brown2020language}, GPT-4 \cite{openai2023gpt4}, Gemini \cite{GeminiReport2023}, and  Llama-2 \cite{touvron2023llama} have been pre-trained on vast and diverse datasets across multiple domains. This pre-training equips them with the remarkable ability to perform natural language processing tasks with minimal or even zero additional training, thus lowering the technological barriers to creating innovative educational solutions. The recent introduction of ChatGPT and Google's BARD marks a significant step towards user-friendly, LLM-based generative chatbots. These user-friendly interfaces enable a broader audience to harness the power of sophisticated language models, contributing to increased accessibility and engagement with artificial intelligence.

Researchers have measured the capability of LLMs to pass specific exams, but primarily to measure the LLMs' power to mimic human intelligence. Mainly, GPTs and BARD models have been tested on a wide range of fields such as the United States Medical Licensing Exam (USMLE) \cite{10.1371/journal.pdig.0000198},  the American Board of Anesthesiology (ABA) exam \cite{Angel2023-cq}, and in a vast datasets in Medicine \cite{fi15060192}; proficiency in reading comprehension \cite{deWinter2023}, and various branches of knowledge, including subjects in the humanities, social sciences, physics, computer science, mathematics, and more \cite{hendryckstest2021},  mainly in English language. Also, for the Spanish language, some studies have been conducted in the medical context, such as the Spanish Medical Intern Examination (MIR) \cite{clinpract13060130}, and Rheumatology-related questions from MIR \cite{Madrid-Garcia2023}.

While the potential benefits of integrating LLMs into education are evident, educators are concerned that the widespread use of LLMs may lead students to overly depend on technology for acquiring factual information and reasoning. They are concerned that students might stop developing their critical thinking skills if they become accustomed to relying solely on LLMs for answers without reasoning. Moreover, educators are apprehensive about the potential for cheating in online exams, where students could exploit LLMs to obtain answers, generate essays, or provide explanations \cite{Cotton2023}.

This study centers on the evaluation of models that offer free accessibility to the majority of Mexican students. We specifically examine two LLMs, GPT-3.5 and BARD (supported by Gemini Pro). Our primary objective is to assess the general knowledge, problem-solving, and reasoning capabilities of GPT-3.5 and BARD. To achieve this, we analyze their performance on three sample exams for undergraduate admissions. Knowledge tests play a crucial role in selecting candidate students equipped with the necessary knowledge to pursue academic programs in biological and medical sciences, engineering, mathematical and physical sciences, and social and administrative sciences.

\section{Material and Methods}
\label{sec:methods}

The National Polytechnic Institute\footnote{https://www.ipn.mx} (IPN, for its acronym in Spanish) is a public institution dedicated to advancing education, research, and innovation. As one of the leading educational institutions in Mexico, holding the estimated rank of the third-best university in the country\footnote{https://edurank.org/geo/mx/}, the IPN plays a pivotal role in providing high-quality academic programs across various disciplines. 

The IPN offers 69 academic programs in three main fields of study: 43 programs in engineering, including mathematics and physics; 14 programs in biological and medical sciences; and 12 programs in social sciences. The IPN publishes a study guide for the university admissions tests each year. The 2023 admissions tests were structured by areas of knowledge. This year, history and reading comprehension of the English language were included to enhance the comprehensive academic program \cite{IPN2023}. The admission exam comprises 140 questions covering subjects such as mathematics, history, writing, and reading skills in the Spanish language, biology, chemistry, physics, and reading comprehension of English as a foreign language.

The admission exams were prepared for three main groupings of fields of study: Engineering/Mathematical and Physical Sciences (E-MPS), Biological and Medical Sciences (BMS), and Social and Administrative Sciences (SAS). Each exam evaluates various vital skills and competencies of candidate students in their field of study. These exams aim to provide a standardized measure of a student's readiness and ability to understand and analyze written passages in both Spanish and English with a deep understanding of Spanish, which includes comprehension, interpretation, and application of information. The physics, chemistry, and math sections assess student's quantitative reasoning, problem-solving, and mathematical knowledge, while the questions also evaluate critical thinking and logical reasoning.

The distribution of exam questions by topics for undergraduate admissions to the three groups of fields of study offered by the IPN academic programs is presented in Table \ref{table:exam_distribution}. The sample exams consist of 140 multiple-choice questions (indicated in the Q column of the table) with varying distribution based on the major chosen by the candidate student. For example, in the case of an engineering career, such as the E-MPS exam, mathematics and physics carry more weight, with 37 and 17 questions, respectively, in contrast to biological (BMS exam) or social (SAS exam) sciences.

On the one hand, LLMs demonstrate exceptional dexterity in processing and interpreting the text. However, not all LLMs used in our experiments can handle visual information. Therefore, we aim to minimize the inclusion of visual questions to ensure a fair comparison. In our experiments, we refrain from using visual information, which includes questions or options involving images such as sequences of figures, schemes, charts, and electrical diagrams. If a question originally included images and could be adequately described in text form, we included the question by providing a textual description of the image.

The distribution of questions prepared and adapted for the experiments is shown in Table \ref{table:exam_distribution}. The column EQ represents the number of questions used in our experiments. The E-MPS and SAS exams consist of 126 questions each, and the BMS exam of 122 questions.

\begin{table}[ht]
\renewcommand{\arraystretch}{1.3}
\centering
 \caption{Exam's question distribution by topics: Engineering/Mathematical and Physical Sciences (E-MPS), Biological and Medical Sciences (BMS), and Social and Administrative Sciences (SAS). EQ is the number of questions used in the assessment of LLMs; Q is the number of questions in the sample exam}
 \resizebox{.7\textwidth}{!}{
	\begin{tabular}{p{0.35\textwidth}rrr}
	\hline
		   & E-MPS &  BMS & SAS  \\
        Topic & EQ/Q  & EQ/Q & EQ/Q \\
	\hline

        Biology  &  8/9 &  17/17 &  10/10  \\
        Chemistry  &  13/17 &  14/17 &  7/10  \\
        Foreign Language   &  9/10 &  9/10 &  9/10  \\
        History  &  10/10 &  10/10 &  19/20  \\
        Mathematics  &  32/37 &  31/33 &  30/35  \\
        Physics  &  17/17 &  11/13 &  8/10  \\
        Reading Comprehension  &  18/20 &  10/20 &  19/20  \\
        Writing Comprehension  &  19/20 &  20/20 &  24/25  \\
        \cline{2-4}
        Total  &  126/140 &  122/140 &  126/140  \\

            	\hline
	\end{tabular}
 }
	\label{table:exam_distribution}
\end{table}

The language models employed for the experiments were GPT-3.5 and BARD. For the GPT-3.5 model, we utilized the ChatGPT web interface \cite{ChatGPT2023} with the version of November 2023, which includes support for the Spanish language. Similarly, to assess BARD, we employed the BARD  web interface \cite{Bard2023} with the updated version of December 2023, supporting the Spanish language and enhances introduced by the Gemini Pro model \cite{Gemini2023} \cite{GeminiReport2023}.

We proceeded manually with the assessment of the models by entering questions along with the corresponding multiple-choice options in the models' web interface. All questions have four response choices (a, b, c, and d). The responses generated by the models were compared to the correct answer sheet for each exam included in the study guide for admissions. 

Sometimes, the model did not respond because the question was not understandable. In such cases, the question was paraphrased or provided with further clarification until the model obtained a response. In addition, we introduced one the following prompts followed by the question to push the model to select an option: \textit{``Seleccionar de las siguientes opciones ...''} (``Select from the following options'') or \textit{``Seleccionar una opción de las siguientes opciones ...''} (``Select an option from the following choices''). 
In the case of the reading and writing section, for text-dependent questions, the text is provided to the model; subsequently, a prompt is used such as  \textit{``Dado el texto anterior,''} (``Given the previous text,'') or \textit{"Del texto anterior,"} (``From the previous text,'') followed by the question along with its multiple choices. 
In the mathematics section, if a question or an option involves mathematical notation, it is represented with its Wolfram form \cite{Wolfram2023}. This approach ensures that both models interpret formulas accurately. The repository containing the questions and their corresponding answers used in the experiments is available for download on GitHub\footnote{https://github.com/sabinomi/Exams4LLMs}.

IPN admissions to an academic program is contingent upon achieving a minimum number of correct answers. The specific number required varies depending on the academic unit and program. Table \ref{table:minimum_scores} summarizes the minimum scores necessary for admitting a candidate student to the school and campus. The scores are provided by the IPN Office of Transparency and Access to Information \cite{ResultsIPN2022}. The columns \textit{Estimated Minimum Score (2023)} and \textit{Estimated Minimum Score} represent the minimum score proportion relative to the minimum score in the year 2022. The admissions exams of the year 2022 comprise 130 questions. The table presents the highest, median, and lowest required minimum values for each of the fields of study.

For example, to be accepted into Aeronautical Engineering (Ingeniería Aeronáutica) at the ESIME Ticomán campus, a minimum score (2022) of 96 correct answers is required, and for the Geophysical Engineering (Ingeniería Geofísica) program at the ESIA Ticomán campus, a score of 70 correct answers is required. Both academic programs belong to the Engineering/Mathematical and Physical Sciences (E-MPS).

In our experiments, we considered the \textit{Estimated Minimum Score} as the required minimum value for acceptance to the campus or the equivalent percentage of the minimum score. 

\begin{table}[ht]
\renewcommand{\arraystretch}{1.3}
\centering
 \caption{Summary of academic programs and the minimum scores required for IPN admissions categorized by fields of study and academic programs: Engineering/Mathematical and Physical Sciences (E-MPS), Biological and Medical Sciences (BMS), and Social and Administrative Sciences (SAS). The term \textit{Minimum Score (2022)} refers to the minimum score mandated by each academic program for student acceptance to the school and campus. The \textit{Estimated Minimum Score} represents the proportion of the minimum score, considering the questions in the experiments based on the minimum score of 2022}
 \resizebox{.99\textwidth}{!}{

        \begin{tabular}{lll>{\centering\arraybackslash}p{2.5cm}>{\centering\arraybackslash}p{2.5cm}>{\centering\arraybackslash}p{2.5cm}} 
         \hline	
        Fields of Study & Academic Program & School/Campus & Minimum Score (2022) & Estimated Minimum Score (2023) & Estimated Minimum Score\\ 
         \hline 
        E-MPS & Ingeniería Aeronáutica & ESIME Ticomán & 96 & 103.4 & 93.0\\
        E-MPS & Ingeniería Biónica & UPIITA & 95 & 102.3 & 92.1\\
        E-MPS & Licenciatura en Física y Matemáticas & ESFM & 90 & 96.9 & 87.2\\
        E-MPS & Ingeniería en Inteligencia Artificial & ESCOM & 90 & 96.9 & 87.2\\
        E-MPS & Ingeniería en Comunicaciones y Electrónica & ESIME Zacatenco & 73 & 78.6 & 70.8\\
        E-MPS & Ingeniería Geofísica & ESIA Ticomán & 70 & 75.4 & 67.8\\
        BMS & Médico Cirujano y Partero & ESM & 98 & 105.5 & 92.0\\
        BMS & Licenciatura en Odontología & CICS Santo Tomás & 97 & 104.5 & 91.0\\
        BMS & Licenciatura en Biología & ENCB & 88 & 94.8 & 82.6\\
        BMS & Licenciatura en Enfermería & ESEO & 83 & 89.4 & 77.9\\
        BMS & Licenciatura en Trabajo Social & CICS Milpa Alta & 72 & 77.5 & 67.6\\
        BMS & Licenciatura en Optometría & CICS Santo Tomás & 72 & 77.5 & 67.6\\
        SAS & Licenciatura en Administración y Desarrollo Empresarial & ESCA Santo Tomás & 98 & 105.5 & 95.0\\
        SAS & Licenciatura en Negocios Internacionales & ESCA Santo Tomás & 93 & 100.2 & 90.1\\
        SAS & Licenciatura en Economía & ESEO & 80 & 86.2 & 77.5\\
        SAS & Contador Público & ESCA Tepepan & 79 & 85.1 & 76.6\\
        SAS & Licenciatura en Turismo & EST & 71 & 76.5 & 68.8\\
        SAS & Licenciatura en Archivonomía & ENBA & 70 & 75.4 & 67.8\\
        \hline
	\end{tabular}
 }
	\label{table:minimum_scores}
\end{table}

\section{Results}

The overall results of the LLMs evaluation are presented in Table \ref{table:overall_scores}. For the Engineering/Mathematical and Physical Sciences exam (E-MPS), GPT-3.5 and BARD achieved an identical score of 57.93\% 
Regarding the Biological and Medical Sciences exam (BMS), BARD performed slightly better with a score of 59.83\%  compared to GPT-3.5. For the Social and Administrative Sciences exam (SAS), GPT-3.5 outperformed BARD with scores of 65.87\% and 63.49\%, respectively, achieving a more successful outcome in the examination. In summary, GPT-3.5 outperformed BARD marginally, securing scores of 60.94\% and 60.42\%, respectively. Considering the minimum acceptance score for the year 2022, which is a percent score of 53.85\% for the E-MPS and SAS exams and a score of 55.38\% for the BMS exam, both models demonstrated sufficient performance for IPN admission to an academic program.

\begin{table}[ht]
\renewcommand{\arraystretch}{1.3}
\centering
 \caption{Overall performance results of the LLMs evaluated on the sample exams}
 \resizebox{.7\textwidth}{!}{
	\begin{tabular}{p{0.1\textwidth}ccc}
	\hline
		Exam & Model & Raw Score & Percent Score \\
	\hline
        \multirow{2}{*}{E-MPS}  &  GPT-3.5   &  73/126   &  57.93 \\
                            &  BARD    &    73/126   &  57.93  \\

        \hline
        \multirow{2}{*}{BMS}  &    GPT-3.5  &  72/122   & 59.01 \\
                              &  BARD   &     73/122      & 59.83  \\
        \hline
        \multirow{2}{*}{SAS} &  GPT-3.5  &   83/126  &  65.87 \\
                            &  BARD &   80/126  &  63.49 \\
	\hline
       \multirow{2}{*}{Average}  & GPT-3.5  & - & \textbf{60.94} \\
                                 & BARD  &  - & 60.42    \\
        \hline
	\end{tabular}
 }
	\label{table:overall_scores}
\end{table}

Tables \ref{table:E-MPS_scores}, \ref{table:BMS_scores}, and \ref{table:SAS_scores} show the disaggregated responses of the exams by topic: Biology, Chemistry, Foreign Language, History, Mathematics, Physics, Reading Comprehension, and Writing Comprehension.  

Table \ref{table:E-MPS_scores} shows the results of the E-MPS exam. The exam consists of 126 questions, covering more questions in Mathematics and Physics. Both models performed well on most topics, and overall performance is identical score of 57.93\%. In the topic-specific performance, GPT-3.5 outperforms BARD in Biology, History, Mathematics, and Writing Comprehension. On the other hand, BARD scored the same as GPT-3.5 in Chemistry and Physics; BARD is better in Foreign Language and Reading Comprehension.

Table \ref{table:BMS_scores} shows the topic-specific performance for the BMS exam, which comprises 122 questions. The exam covers more questions for Biology and Chemistry. Overall performance, BARD outperforms slightly GPT-3.5 on the BMS exam, with a score of 59.83\% compared to GPT-3.5 of 59.01\%. BARD performed better than GPT-3.5 on all topics except for Mathematics and Physics. The most significant difference in performance was in Physics, where BARD scored 36.36\% compared to GPT-3.5, with a score of 72.73\%.

Table \ref{table:SAS_scores} shows the topic-specific performance for the SAS exam, which comprises 126 questions. The exam covers more questions for History, Reading Comprehension, and Writing Comprehension, which covers 49.21\% of the exam. Overall performance, GPT-3.5 outperforms BARD on the SAS exam, with a score of 65.87\% compared to BARD of 63.49\%. GPT-3.5 performed better than BARD in Mathematics, Physics, Reading, and Writing Comprehension. BARD does better in Biology, Foreign Language, and History. Both models performed well in Chemistry.

\begin{table}[ht]
\renewcommand{\arraystretch}{1.3}
\centering
 \caption{Results by topics, Engineering/Mathematical and Physical Sciences exam (E-MPS). CA = correct answers by the model, Q = total questions in the topic}
 \resizebox{.9\textwidth}{!}{
	\begin{tabular}{p{0.35\textwidth}cccc}
         \hline	
         \multirow{2}{*}{Topic}  & \multicolumn{2}{c}{GPT-3.5} & \multicolumn{2}{c}{BARD} \\
         \cline{2-5} 
         & CA/Q & Percent Score & CA/Q & Percent Score\\ 
         
         \hline 
            Biology  &   7/8  &   87.5  &   6/8  &   75.0\\
            Chemistry  &   6/13  &   46.15  &   6/13  &   46.15\\
            Foreign Language   &   6/9  &   66.67  &   9/9  &   100.0\\
            History  &   7/10  &   70.0  &   6/10  &   60.0\\
            Mathematics  &   16/32  &   50.0  &   13/32  &   40.62\\
            Physics  &   12/17  &   70.59  &   12/17  &   70.59\\
            Reading Comprehension  &   8/18  &   44.44  &   11/18  &   61.11\\
            Writing Comprehension  &   11/19  &   57.89  &   10/19  &   52.63\\        
         \hline 

	\end{tabular}
 }
	\label{table:E-MPS_scores}
\end{table}

\begin{table}[ht]
\renewcommand{\arraystretch}{1.3}
\centering
 \caption{Results by topics, Biological and Medical Sciences Exam (BMS). CA = correct answers by the model, Q = total questions in the topic}
 \resizebox{.9\textwidth}{!}{
	\begin{tabular}{p{0.35\textwidth}cccc}
         \hline	
         \multirow{2}{*}{Topic}   & \multicolumn{2}{c}{GPT-3.5} & \multicolumn{2}{c}{BARD} \\
         \cline{2-5} 
         & CA/Q & Percent Score & CA/Q & Percent Score\\ 
         \hline 
        Biology  &   10/17  &   58.82  &   12/17  &   70.59\\
        Chemistry  &   7/14  &   50.0  &   8/14  &   57.14\\
        Foreign Language   &   6/9  &   66.67  &   7/9  &   77.78\\
        History  &   6/10  &   60.0  &   8/10  &   80.0\\
        Mathematics  &   17/31  &   54.84  &   14/31  &   45.16\\
        Physics  &   8/11  &   72.73  &   4/11  &   36.36\\
        Reading Comprehension  &   5/10  &   50.0  &   6/10  &   60.0\\
        Writing Comprehension  &   13/20  &   65.0  &   14/20  &   70.0\\        
        \hline
	\end{tabular}
 }
	\label{table:BMS_scores}
\end{table}

\begin{table}[ht]
\renewcommand{\arraystretch}{1.3}
\centering
 \caption{Results by topics,  Social and Administrative Sciences exam (SAS).  CA = correct answers by the model, Q = total questions in the topic}
 \resizebox{.9\textwidth}{!}{
	\begin{tabular}{p{0.35\textwidth}cccc}
         \hline	
         \multirow{2}{*}{Topic}   & \multicolumn{2}{c}{GPT-3.5} & \multicolumn{2}{c}{BARD} \\
         \cline{2-5} 
         & CA/Q & Percent Score & CA/Q & Percent Score\\ 
         \hline 
        Biology  &   6/10  &   60.0  &   10/10  &   100.0\\
        Chemistry  &   7/7  &   100.0  &   7/7  &   100.0\\
        Foreign Language   &   6/9  &   66.67  &   7/9  &   77.78\\
        History  &   14/19  &   73.68  &   16/19  &   84.21\\
        Mathematics  &   19/30  &   63.33  &   16/30  &   53.33\\
        Physics  &   6/8  &   75.0  &   4/8  &   50.0\\
        Reading Comprehension  &   11/19  &   57.89  &   9/19  &   47.37\\
        Writing Comprehension  &   14/24  &   58.33  &   11/24  &   45.83\\
        \hline
	\end{tabular}
 }
	\label{table:SAS_scores}
\end{table}

\begin{figure}[ht]
\centering
\includegraphics[width=0.8\textwidth]{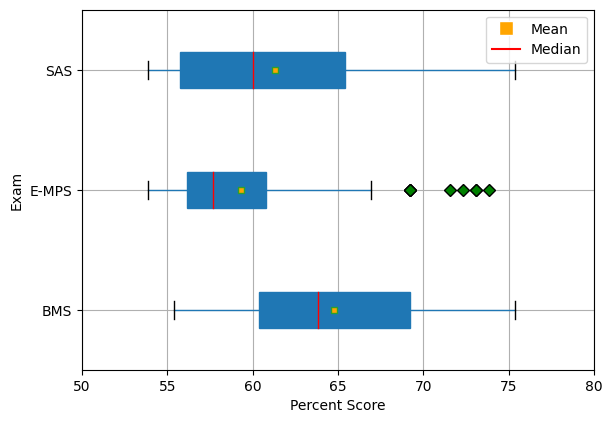}
\caption{Quartiles of required minimum percent scores for the academic programs offered by the IPN covering the three main groups of fields of study.}
\label{fig:estimated_scores}
\end{figure}

Figure \ref{fig:estimated_scores} illustrates the quartiles of required minimum percentage scores for academic programs offered by the IPN, encompassing three main groups of fields of study. For the E-MPS exam, the quartiles are Q1 = 56.15, Q2 = 57.69, and Q3 = 60.77, with a minimum value of 53.85 and a maximum value of 73.85. Both GPT-3.5 and BARD perform slightly higher (57.93) than Q2. Consequently, the models have achieved admission to at least 50\% of the schools/campuses offering academic programs in Engineering and Mathematical and Physical Sciences.

For the BMS exam, the quartiles are Q1 = 60.39, Q2 = 63.85, and Q3 = 69.23, with a minimum score of 55.38 and a maximum score of 75.38. GPT-3.5 and BARD, with respective scores of 59.01 and 59.83, meet the criteria for admission to 25\% of schools offering academic programs in Biological and Medical Sciences.

In the case of the SAS exam, the quartiles are Q1 = 55.77, Q2 = 60.0, and Q3 = 65.39, with a minimum score of 53.85 and a maximum score of 75.38. GPT-3.5 slightly exceeds Q3, representing the minimum score required for admission to 75\% of academic programs in the Social and Administrative Sciences field. BARD scored 63.49, placing it between the 50\%-75\% acceptance range in the same field of study.

\section{Discussion}

Although the required minimum score varies from year to year, the percentage of the minimum score is an excellent measure to estimate the proficiency of the models.
According to the results, the evaluated models have exhibited proficiency in successfully passing all three admission exams. 
GPT-3.5 consistently outperforms BARD in Mathematics across all exams. However, the overall performance in this subject remains relatively low, less than 63.33\%. Furthermore, GPT-3.5 outperforms BARD in Physics in two exams (E-MPS and BMS). These subjects involve tasks that require comprehension, reasoning, problem-solving, and calculation. Exam questions cover diverse topics such as numerical series, geometry problems, systems of linear equations, trigonometry problems, and differential and integral calculus.

The models excel in solving specific and well-known academic problems where an apparent problem is presented, and a formula can be applied to find a solution, such as questions related to calculus or systems of linear equations. However, the models encounter challenges when faced with math word problems presented in a textual format requiring students to interpret and solve the problem. In such situations, GPT-3.5 and BARD may need help because the problem statement and sequence of explanations are well-defined; calculating or substituting values may pose difficulties. In these cases, a second interaction with the model was initiated solely for examination purposes to indicate the error in the model's response and provide the correct information. The model then adjusted its selected choice and explained the solution but often encountered further inaccuracies.

Evaluating the overall performance in Spanish language-related questions, encompassing reading and writing comprehension topics across all exams, GPT-3.5 achieved a percentage score of 56.36\%, while BARD attained a score of 55.45\%. These scores correspond to raw scores of 62/110 and 61/110 questions for GPT-3.5 and BARD, respectively. Both models encountered challenges in identifying the text's central ideas, determining point of view and tone, and engaging in textual entailment to infer information. Notably, BARD exhibited proficiency in tasks requiring factual information, such as history-related or conceptual problems.

\section{Conclusions}
\label{sec:conclusions}

LLMs have demonstrated proficiency in successfully passing all three exams required for IPN undergraduate admissions in the Spanish language. The models' notable achievements enable admission to up to 75\% for some academic programs. 
However, the most sought-after academic programs, representing the top 25\%, such as Medical Doctor and Obstetrician, Aeronautical Engineering, Business Administration and Development, Artificial Intelligence Engineering, and Bachelor's in Physics and Mathematics, among others, currently fall beyond the scope of these models.

Due to LLMs becoming widely used, it may be necessary to modify the format and content of exams to ensure fair and reliable assessments for all students. 
The widespread availability of advanced LLMs could create an unfair advantage for some students, exacerbating existing educational inequalities and placing underprivileged students at a further disadvantage.

Despite the challenges LLMs pose, they also present promising opportunities for education. These models have demonstrated strong capabilities in supporting the learning process through detailed explanations during problem-solving and the ability to refine answers through interactions. However, it is crucial to note that, for now, these models are not entirely reliable.


\bibliographystyle{plain}
\bibliography{biblio}


\end{document}